%% file: 0-main.tex
\pdfoutput=1

\documentclass[11pt]{article}

\usepackage{EMNLP2023}

\usepackage{times}
\usepackage{latexsym}
\usepackage{multirow}

\usepackage[T1]{fontenc}

\usepackage[utf8]{inputenc}

\usepackage{microtype}

\usepackage{inconsolata}

\title{How Can Context Help? \\  Exploring Joint Retrieval of Passage and Personalized Context}

\author{Hui Wan \\
  IBM Research AI \\
  \texttt{hwan@us.ibm.com} \\\And
  Hongkang Li \\
  Rensselaer Polytechnic Institute \\
  \texttt{lih35@rpi.edu} \\\AND
  Songtao Lu \\
  IBM Research AI \\
  \texttt{songtao@ibm.com} \\\And
  Xiaodong Cui \\
  IBM Research AI \\
  \texttt{cuix@us.ibm.com} \\\And
  Marina Danilevsky \\
  IBM Research AI \\
  \texttt{mdanile@us.ibm.com} \\
  }

\usepackage[colorinlistoftodos,prependcaption,textsize=tiny]{todonotes}
\usepackage{amsmath}
\usepackage{enumitem}

\begin{document}
\maketitle
\begin{abstract}

The integration of external personalized context information into document-grounded conversational systems has significant potential business value, but has not been well-studied. Motivated by the concept of personalized context-aware document-grounded conversational systems, we introduce the task of context-aware passage retrieval. We also construct a dataset specifically curated for this purpose. We describe multiple baseline systems to address this task, and propose a novel approach, Personalized Context-Aware Search (PCAS), that effectively harnesses contextual information during passage retrieval. Experimental evaluations conducted on multiple popular dense retrieval systems demonstrate that our proposed approach not only outperforms the baselines in retrieving the most relevant passage but also excels at identifying the pertinent context among all the available contexts.  We envision that our contributions will serve as a catalyst for inspiring future research endeavors in this promising direction. 

\end{abstract}

\input{1-intro}

\vspace{20pt}
\input{2-task}

\vspace{20pt}
\input{3-dataset}

\vspace{20pt}
\input{4-approach}

\vspace{20pt}
\input{5-exp}

\vspace{20pt}
\input{6-related}

\vspace{20pt}
\input{7-conclusion}

\vspace{20pt}
\input{98-limit}

\section*{Acknowledgements}
This work was partly supported by the Rensselaer-IBM AI Research Collaboration (\url{http://airc.rpi.edu}), part of the IBM AI Horizons Network (\url{http://ibm.biz/AIHorizons}).
\bibliography{anthology,custom}
\bibliographystyle{acl_natbib}

\input{99-appendix}

\end{document}

%% file: 1-intro.tex
\section{Introduction}
\label{sec-intro}

With the recent developments in AI, the world has witnessed an eruption of chatbots deployed with LLMs (large language models), such as ChatGPT, BARD~\cite{lamda2022}, BlenderBot~\cite{shuster2022blenderbot}, which often generate texts indistinguishable from human fluency. However, chatbots powered by parameter-based LLMs are known to generate factually incorrect statements - a problem regardless of the model size \cite{shuster-etal-2021-retrieval-augmentation}. 
By leveraging an external corpus of knowledge, retrieval augmented systems \cite{wizardofwikipedia, lewis2020retrieval, karpukhin-etal-2020-dense}, including document-grounded dialogue systems \cite{roller2020recipes, shuster2022blenderbot, shuster-etal-2022-language, lamda2022}, have demonstrated several advantages compared to pure parameter-based systems. For instance, grounding responses on external knowledge bases have been shown to reduce hallucinations across a variety of retrieval systems and model architectures \cite{shuster-etal-2021-retrieval-augmentation}.

A document-grounded conversational system, particularly in the enterprise setting, is likely to have access to a significant amount of contextual information, whether as a knowledge base or a library of API calls. This context may be temporal, such as the current date and time, or recent events; or it may be user-specific, such as information about the user's account, profile, recent transactions, activity logs, etc. Without any such context, a user's question such as, "Am I eligible for this rebate?" would receive the generic answer "You may be eligible for this rebate depending on where you live," grounded only on the relevant documents. If the right context were also retrieved and made available, the response could be instantly elevated to "Yes, you are eligible since you live in Singapore." Furthermore, retrieving the correct context information may serve to better understand the user's intent, and therefore improve the likelihood of identifying the correct grounding document to use. 

The significant challenge of choosing which context to retrieve has great potential business value, but has not been well-studied. Including too much contextual information may result in too much noise to the generation step, or exceeding the LLM's allowed input size. Including irrelevant contextual information  may degrade the generated response. Our motivating question can thus be posed as follows: Given a user query (which itself may be underspecified), a document collection, and a set of available contexts, how can a document-grounded conversational system retrieve a good subset of contexts to help answer the query, and can this process also help in retrieving the most relevant grounding document(s)?

To this end, we propose a new task of personalized context-aware passage retrieval for document-grounded dialogue,
and create a dataset, ORCA-ShARC, for this setting. 
We provide several baseline approaches, as well as develop a novel approach, Personalized Context-Aware Search (PCAS), to address the task. 

In order to showcase the efficacy of PCAS, we conduct extensive experiments on multiple well-known retrieval systems. The results illustrate that PCAS not only surpasses the baselines in retrieving the most relevant passage but also excels in identifying the pertinent context. 
We hope that our advancements in joint context-passage retrieval will serve as a catalyst, motivating future research endeavors in this highly promising field.



%% file: 2-task.tex
\vspace{-0.5cm}

\section{Context-and-Passage Retrieval}
\label{sec-task}
We now formally define the task of context-passage retrieval, which involves not only retrieving the relevant document from the external knowledge base, but also selecting the relevant piece of context from all the available contexts.

Formally, when a user $u$ engages in a conversation with the system, in addition to the static document corpus $\mathcal{D}$, all the available personalized context information $\mathcal{C}^u$ are accessible to the system.
There is also the conversation history composed of utterances that have already occurred in this session between the user and the system: $H=\{r_1:X_1, ..., r_{i}:X_{i}, ...\}$ where $r_i$ is the speaker role, and $X_i$ is the utterances at the $i$-th turn, respectively.
Since the focus of our work is in context-passage retrieval rather than the conversation history, in the rest of the paper, we simply consider a single turn user query $q$ rather than
$H$.


Given the input of a user query $q$, 
the task is to select 1) the most relevant latent document 
$d$
from $\mathcal{D}$;
and 2) the most relevant latent  context $c$
from $\mathcal{C}^u$, to help the system generate a good response.


To evaluate the retrieved documents and context, we use standard retrieval metrics, including binary rank-aware  metrics MAP (mean average precision) and decision support metrics Recall@$K$.




%% file: 3-dataset.tex
\vspace{-0.2cm}

\section{The ORCA-ShARC Dataset}



To the best of our knowledge, there is no existing open-retrieval content-grounded dialogue or QA datasets where each document-grounded example is annotated with a set of context. To this end, we curate a dataset for the proposed task in Section~\ref{sec-task}.


ShARC~\cite{saeidi-etal-2018-interpretation} is a conversational QA dataset focusing on question answering from given text and one piece of given context (scenario).
OR-ShARC~\cite{gao2021orsharc} is adapted from the ShARC dataset to an open-retrieval setting, where the task is to retrieve the relevant text snippet from the whole corpus. In OR-ShARC, each example is given one piece of relevant context (scenario).

We create a dataset, ORCA-ShARC (Open-Retrieval Context-Aware ShARC), by converting the OR-ShARC dataset into our task setting, where a set of contexts is provided for each example.
To create the set, we use the example's original relevant context, and expand the set by randomly sampling from all the contexts appearing in the OR-ShARC dataset, as long as there is no contradiction between contexts introduced (as judged by prompting \texttt{FLAN\_T5\_3B} model~\cite{https://doi.org/10.48550/arxiv.2210.11416}).  We include 10 pieces of context for each example.\footnote{As the size of the context set grows, it naturally becomes harder to add context without contradictions. A few examples could only support 6-9 pieces of context.}
Table \ref{tab:dataset} summarizes the statistics of the ORCA-ShARC dataset and Table \ref{tab:example} provides an example.



\begin{table}[ht]
\centering
\small
\begin{tabular}{lc}
\hline\hline
\# Documents (Rule Text Snippets) & 651    \\
\# Avg. Document Length & 38.5  \\
\# Avg. Pieces of Context & 9.94 \\
\# Training Examples & 17936 \\
\# Validation Examples & 1105 \\
\# Test Examples & 2373 \\
\hline\hline
\end{tabular}
\caption{Summary statistics of ORCA-ShARC.}
\label{tab:dataset}
\end{table}


\begin{table}[ht]
\centering
\small
\begin{tabular}{p{0.7in}|p{2in}}
\hline\hline
Source URL &  \url{https://www.gov.uk/winter-fuel-payment/eligibility}  \\
Scenario Set &  ...\\
 & I am and have been an eligible veteran. \\
 & I live in the Swiss Alps.\\
 & I'm trying to export some boots. \\
 & ... \\
Question & Can I get Winter Fuel Payment? \\
Gold Snippet & ...you might still get the payment if both the following apply: * you live in Switzerland or a European Economic Area (EEA) country...  \\
Gold Scenario & I live in the Swiss Alps. \\
\hline\hline
\end{tabular}
\caption{An example from ORCA-ShARC. Note how the Scenario Set is expanded from the one piece of gold Scenario originally annotated in OR-ShARC.}
\label{tab:example}
\end{table}

%% file: 4-approach.tex
\section{Approach}\label{sec-approach}
We compare our approach with three baselines, and use some of the most popular neural retrieval systems to address the context-and-passage retrieval task on the newly constructed dataset.

\subsection{Baselines}
We design and implement several baselines for the task. 
The approaches are independent of the underlying retrieval systems.
We use $score_{dq}(d, q)$, $score_{cq}(c, q)$ and $score_{cd}(c, d)$ to represent the scores from the retrievers to model the pairwise relevance of document $d$, query $q$, and context $c$.


\begin{description}[style=unboxed,leftmargin=0cm]
\item[OR] \{question + original relevant context\} $\xrightarrow{}$ document:
For clarification, this is an experiment on the original OR-ShARC dataset as a reference rather than a baseline. Knowing the original relevant context, the passage retrieval task in OR-ShARC is easier than our task. In this experiment, the original context is concatenated with the user question, forming a new query $q^{\textrm{OR}}$ to retrieve documents based on $score_{dq}(d, q^{\textrm{OR}})$.

\item[B1] \{question + all contexts\} $\xrightarrow{}$ document:
A baseline that concatenates the user question together with all available contexts to form a new query $q^{\textrm{B1}}$ and retrieve documents based on $score_{dq}(d, q^{\textrm{B1}})$.

\item[B2] question $\xrightarrow{}$ document; document $\xrightarrow{}$ context:
A baseline that uses the user question to retrieve documents based on $score_{dq}(d, q)$, then uses the top predicted documents to select contexts based on $score_{cd}(c, d)$. 

\item[B3] question $\xrightarrow{}$ context; \{question + predicted context\} $\xrightarrow{}$ document:
A baseline that uses the user question to select contexts based on $score_{cq}(c, q)$, then concatenates the user question with the top-1 predicted context to form a new query $q^{\textrm{B3}}$ and retrieves documents based on $score_{dq}(d, q^{\textrm{B3}})$.

\end{description}

\subsection{PCAS Approach}\label{subsec:PCAS}
We propose a novel approach, PCAS, that jointly 
predicts the document and the context as a pair,
based on the relevance of the document to both the query and the context.

First, we use the user question $q$ to retrieve the top $K$ document candidates based on $score_{dq}(d, q)$.
Then, for each document $d$, we select the context that is most relevant to it based on $score_{dc}(d, c)$.
Last but not least, a convex combination score 
$\lambda * score_{dq}(d, q) + (1-\lambda) * score_{dc}(d, c)$ is used to select the most relevant pair $(d, c)$ where $0<\lambda<1$.
The underlying intuition is as follows: the user question might not contain sufficient information 
for the system to understand the intent and retrieve the gold document. However, the system will partially know the intent, and has a good chance of including the best document in the top-$K$ list. Matching the top-$K$ documents with the user's actual situation, which is captured in the contexts, will greatly help decipher the user's true intent and retrieve the gold document.


%% file: 5-exp.tex
\vspace{-0.5cm}

\section{Experimental Results}\label{sec-experiments}


We evaluate the baselines and PCAS 
in a $0$-shot context-and-passage retrieval task on the ORCA-ShARC dataset.
We conduct experiments using several popular pretrained modern neural retrieval systems
including a late-interaction retrieval model ColBERT~\cite{khattab2020colbert, santhanam2021colbertv2}, single-vector retrieval models DPR~\cite{karpukhin-etal-2020-dense}, ANCE~\cite{xiong2021ance} and Sentence BERT (S-BERT) \cite{reimers-2019-sentence-bert} with DistilBERT-TAS-B model~\cite{sanh2019distilbert, Hofstaetter2021_tasb_dense_retrieval} .

For ColBERT, we adapt the code from the ColBERT repository
\footnote{\url{https://github.com/stanford-futuredata/ColBERT}}.
From the BEIR \cite{thakur2021beir} repository\footnote{\url{https://github.com/beir-cellar/beir}}, we get the pretrained model names, as well as the code 
for the other dense retrieval systems.
 We use the py\_trec tool\cite{VanGysel2018pytreceval} \footnote{\url{https://github.com/cvangysel/pytrec_eval}} for evaluation.


 In Table \ref{tab:result-doc}, we present the document retrieval results and the context selection results (limited to the approaches that predict the context: B2, B3 and PCAS). 
 For the same approach, the results largely varies across the retrievers, due to the distinct models and different pre-training data and processes. For example, DPR results are on the low side, especially for B3 document metrics, because it involves two chained retrieval steps that amplify this effect.
  B1 yields the lowest accuracy across the board, mainly due to the noise introduced by including all the contexts without discrimination.
 Importantly, we observe that
 when the original relevant context is unknown, 
 our proposed PCAS approach achieves better retrieval results than all baselines, which indicates that jointly considering the documents and contexts can improve the performances of both document and context retrieval. Note that the PCAS results are close to the OR experiment in which the original relevant context is given. This suggests that PCAS can identify the relevant and important contexts for the query, with no need for users to specify any contexts. Furthermore, the comparison between B2 and B3 illustrates that the retrieval process of $q\rightarrow d\rightarrow c$ is better than $q\rightarrow c\rightarrow d$ , which supports the motivation of our PCAS design.
\begin{table}[ht]
\centering
\small
\begin{tabular}{lc|ccc|c}
\hline
\hline
\multicolumn{2}{c|}{ } & \multicolumn{3}{c|}{documents} & contexts\\
\hline
\multicolumn{2}{c|}{\textbf{Methods}} & R@1 & R@5 & M@5 &R@1 \\
\hline
\hline
\multicolumn{2}{c|}{OR} & 60.81 & 84.25 & 70.52 & NA\\
\hline
\multirow{4}{*}{\rotatebox{90}{ColBERT}}&B1 & 34.93 & 59.82 & 44.47 &NA\\
&B2 & 55.57 & \textbf{90.77} & 70.30 & \textbf{27.24} \\
&B3 & 52.85 & 78.01 & 62.34 &20.90\\
&PCAS & \textbf{59.19} & \textbf{90.77} & \textbf{71.78} & 25.79\\
\hline
\hline
\multicolumn{2}{c|}{OR} & 31.58 & 54.93 & 40.78 & NA\\
\hline
\multirow{4}{*}{\rotatebox{90}{DPR}}& B1 & 14.84 & 35.66 & 22.24 &NA\\
&B2 & 28.96 & 51.67 & 38.38 &32.13\\
&B3 & 18.91 & 38.73 & 25.94 &31.58\\
&PCAS & \textbf{30.04} & \textbf{56.02} & \textbf{38.81} &\textbf{32.49}\\
\hline
\hline
\multicolumn{2}{c}{OR}& 58.37 & 84.52 & 68.69 &NA\\
\hline
\multirow{4}{*}{\rotatebox{90}{ANCE}}&B1 & 41.18 & 73.85 & 53.01 &NA\\
&B2 & 44.98 & 77.29 & 55.35 &37.74\\
&B3 & 42.53 & 75.48 & 55.46 &31.04\\
&PCAS & \textbf{52.85} & \textbf{83.26} & \textbf{65.71} &\textbf{41.36}\\
\hline
\hline
\multicolumn{2}{c}{OR} & 68.15 & 91.40 & 77.55 &NA\\
\hline
\multirow{4}{*}{\rotatebox{90}{S-BERT}} &B1 & 45.43 & 75.84 & 57.00 &NA\\
&B2 & 55.57 & \textbf{91.04} & 68.57 &39.01\\
&B3 & 53.30 & 82.26 & 63.53 &23.44\\
&PCAS & \textbf{63.53} & \textbf{91.04} & \textbf{74.28} &\textbf{42.80}\\
\hline
\hline
\end{tabular}
\caption{Evaluation for document retrieval (first three columns) and context retrieval (last column) on the ORCA-ShARC validation set. R@K denotes Recall@K scores. M@5 denotes MAP@5 scores. NA means that this approach does not retrieve context.}
\label{tab:result-doc}
\end{table}

%% file: 6-related.tex
\vspace{-1cm}

\section{Related Work}
Our proposed task is different from the work on context-aware QA~\cite{seonwoo2020context, taunk2023grapeqa} in which QA is done on a given document and no retrieval is involved. On the other hand, context-aware QA can be considered as the next step following our task.

Our work is closely related to the work on contextual information retrieval~\cite{MERROUNI2019191}.
The major difference from our work is that the tasks in this line of work do not involve selecting of relevant context.
 The form of context being used is structured and in a few pre-defined genres, whereas we leverage a set of unstructured contexts.
There is also no joint relevance modeling of both context and content with respect to the query. 
Our task and approach is also different from contextual recommendation systems (on which \citet{tourani2023capri} recently presents their work), which does not involve user questions or queries.

Our proposed data and task are also related to open domain question answering \citep{kwiatkowski-etal-2019-natural,lewis2020retrieval,min-etal-2020-ambigqa,qu2020open, izacard2021distilling, li2021graph, xiong2021answering, yu2021fewshot},
 open-retrieval conversational QA
\citep{qu2020open, gao2021orsharc}, and open-retrieval document-grounded dialogues~\cite{feng-etal-2021-multidoc2dial}.
However, none of these datasets and tasks include external context information. The lone exception is OR-SHARC, which provides one relevant context for each example and does not involve any selection of relevant context from a larger set. 

Finally, our work is related to Multi-Session Chat (MSC)~\cite{qian2021learning},
a dataset consisting of multiple chat sessions, created for studying how to 
utilize information outside of the current conversation session.
Similarly, ~\citet{xu-etal-2022-beyond} recently leveraged retrieval-augmented methods to select useful contexts from previous chat sessions. 
However, the datasets in both works are not document-grounded, and  retrieving documents jointly with contextual information is not explored.

%% file: 7-conclusion.tex
\vspace{-0.5cm}

\section{Conclusion}
This work proposes the task of context-aware passage retrieval and creates a dataset based on OR-ShARC. We also present a novel approach for integrating external context information into the retrieval aspect of document-grounded conversational systems. The proposed PCAS method effectively combines both the document-query relevance and contextual relevance.
We conduct experimental evaluations on popular retrieval systems, including ColBERT, DPR, ANCE, and S-BERT. The results demonstrate that incorporating external context information through PCAS brings significant improvement on passage retrieval and achieves higher MAP and Recall@$K$ than baseline models. 

The proposed retrieval paradigm opens up avenues for future research and extensions. Several potential directions include extending the PCAS method to the training process, integrating downstream modules such as response generation, and creating real-world context datasets with the inclusion of human feedback. These topics offer promising directions for the community to explore and advance the field further.




%% file: 98-limit.tex
\section*{Limitations}
Although the motivation for our work is to improve the quality of conversation systems that could be grounded on both document content and contextual information, in this work we focus exclusively on the task of retrieving the content and context. We do not evaluate the resulting generative responses to the user query, which is the actual final desired outcome, though we are working on doing so in the near future. 

There are no publicly available datasets that are suited to the context-aware passage retrieval task. 
To the best of our knowledge, some of the related datasets do not have a passage retrieval setting, the rest of them either do not have annotated context set, or have a few limited genre contexts which does not serve our motivation with rich context from various business applications. This leads us to create a dataset by ourselves. Experimenting on only one dataset may limit the more generalizable conclusions we can draw.

The size of the context set that we construct is limited. In a realistic setting, it is likely that the number of available contexts is much greater, and more heterogeneous in nature rather than the short unstructured text available  from the OR-ShARC dataset. More work is needed to discover how our approach scales to these settings.

\section*{Ethics Statement}
To the best of our knowledge, we have not identified any immediate ethical concerns or negative societal consequences arising from this work, in accordance with the ACL ethics policy. The dataset we created does not involve generating any data with LLMs, hence does not impose any risk of non-factual or harmful content.

We hope our work may serve as an inspiration for the community to utilize the newly created dataset and further enhance retrieval performance. In the future, it may be necessary to tread carefully when incorporating user context in a real-world setting, for example to not cause consternation about how much a system may know about a given user. 

%% file: 99-appendix.tex
\clearpage
\appendix

\section{Appendix}
\label{sec:appendix}

\subsection{Experimental setups}

\begin{description}[style=unboxed,leftmargin=0cm]

\item[Models Used] We use the models ``facebook-dpr-question\_encoder-multiset-base''\footnote{\url{https://huggingface.co/facebook/dpr-question_encoder-multiset-base}} and ``facebook-dpr-ctx\_encoder-multiset-base''\footnote{\url{https://huggingface.co/facebook/dpr-ctx_encoder-multiset-base}} for DPR \cite{karpukhin-etal-2020-dense}, ``msmarco-roberta-base-ance-firstp''\footnote{\url{https://huggingface.co/sentence-transformers/msmarco-roberta-base-ance-firstp}} for ANCE \cite{xiong2021ance}, and ``msmarco-distilbert-base-tas-b''\footnote{\url{https://huggingface.co/sentence-transformers/msmarco-distilbert-base-tas-b}} \cite{sanh2019distilbert, Hofstaetter2021_tasb_dense_retrieval} for Sentence BERT \cite{reimers-2019-sentence-bert}. We use the ColBERT model from Beir website \footnote{\url{https://public.ukp.informatik.tu-darmstadt.de/thakur/BEIR/models/ColBERT/msmarco.psg.l2.zip}}.

\item[Hyper-parameters] On the validation set, for DPR, ANCE, and S-BERT, we set $\lambda=0.6$ and $beam=7,\ 7,\ ,5$ for these three systems, respectively. ``$beam$'' means the number of top document candicates retrieved using $score_{dq}(d,q)$ in the first step of the PCAS approach. Please see Section \ref{subsec:PCAS} for more details. On the test set, for DPR, ANCE, and S-BERT, we keep $\lambda=0.6$ and all $beam$'s equal to $5$. For ColBERT, we set $\lambda=0.55$ and $beam$ to $5$ on both test set and validation set.
 
\end{description}
 
\subsection{Results on the test set}
We present evaluation results for document retrieval on the test set with systems in Section \ref{sec-approach} in Table \ref{tab:result-doc-test}
. 
We notice that the performance of the baselines are not stable on the test set across different retrievers. For example, with S-BERT, OR yields much better results than B2, but with ColBERT, B2 performs much better than OR, whereas the difference between B2 and OR in document retrieval is whether to concatenate the given "gold" context to the query. This indicates that there might be noisy data in the test set, and the characteristic of late interaction in ColBERT might be at a disadvantage in the test set data. It also explains why B2 outperforms both OR and PCAS with ColBERT.

\begin{table}[ht]
\centering
\small
\begin{tabular}{lc|ccc|c}
\hline
\hline
\multicolumn{2}{c|}{ } & \multicolumn{3}{c|}{documents} & contexts\\
\hline
\multicolumn{2}{c|}{\textbf{Methods}} & R@1 & R@5 & M@5 & R@1 \\
\hline
\hline
\multicolumn{2}{c|}{OR} & 66.16 & 85.08 & 73.32 &NA \\
\hline
\multirow{4}{*}{\rotatebox{90}{ColBERT}}&B1 & 39.36 & 60.30 & 47.40 &NA\\
&B2 & \textbf{70.50} & \textbf{93.30} &  \textbf{79.09} &\textbf{34.77}\\
&B3 & 58.49 & 78.85 & 66.49 &28.87\\
&PCAS & 67.09 & \textbf{93.30} & 77.07 &32.79 \\
\hline
\hline
\multicolumn{2}{c|}{OR} & 33.21 & 56.38 & 41.70 & NA\\
\hline
\multirow{4}{*}{\rotatebox{90}{DPR}}& B1 & 17.28 & 33.46 & 22.99 &NA\\
&B2 & 29.25 & 57.10 & 39.31 &34.13\\
&B3 & 23.30 & 41.93 & 29.89 &33.59\\
&PCAS & \textbf{29.92} & \textbf{58.96} & \textbf{40.24} &\textbf{35.74}\\
\hline
\hline
\multicolumn{2}{c|}{OR}& 63.55 & 83.52 & 71.20 &NA\\
\hline
\multirow{4}{*}{\rotatebox{90}{ANCE}}&B1 & 49.73 & 72.19 & 58.08 &NA\\
&B2 & 59.80 & \textbf{83.73} & 68.72 &45.51\\
&B3 & 54.45 & 77.24 & 62.95 &36.83\\
&PCAS & \textbf{59.92} & \textbf{83.73} & \textbf{68.94} & \textbf{45.89}\\
\hline
\hline
\multicolumn{2}{c|}{OR} & 69.91 & 89.13 & 77.38 &NA\\
\hline
\multirow{4}{*}{\rotatebox{90}{S-BERT}} &B1 & 47.32 & 75.18 & 57.90 &NA\\
&B2 & 60.94 & \textbf{88.50} & 72.42 &42.98\\
&B3 & 55.29 & 78.13 & 63.85 &28.70\\
&PCAS & \textbf{66.75} & \textbf{88.50} & \textbf{75.29} &\textbf{44.88}\\
\hline
\hline
\end{tabular}
\caption{Evaluation results for document retrieval (first three columns) and context retrieval (last column) on the ORCA-ShARC test set. R@K denotes Recall@K scores. M@5 denotes MAP@5 scores. NA (not applicable) means that this approach does not involve retrieving context.}
\label{tab:result-doc-test}
\end{table}